\documentclass[9 pt]{INTERSPEECH2023}

\interspeechcameraready 

\usepackage{multirow}

\title{PATCorrect: Non-autoregressive Phoneme-augmented Transformer \\
for ASR Error Correction}
\name{Ziji Zhang$^1$, Zhehui Wang$^2$, Rajesh Kamma$^2$, Sharanya Eswaran$^2$, Narayanan Sadagopan$^2$}
\address{
  $^1$Department of Applied Mathematics and Statistics, Stony Brook University, Stony Brook, NY, USA\\
  $^2$Amazon.com Inc, Seattle, WA, USA}
\email{ziji.zhang@stonybrook.edu, zhehwang@umich.edu, \{srikamma,sharanye,sdgpn\}@amazon.com}

\begin{document}

\maketitle
\graphicspath{ {./figs/} }
\begin{abstract}
Speech-to-text errors made by automatic speech recognition (ASR) systems negatively impact downstream models. Error correction models as a post-processing text editing method have been recently developed for refining the ASR outputs. However, efficient models that meet the low latency requirements of industrial grade production systems have not been well studied. We propose PATCorrect-a novel non-autoregressive (NAR) approach based on multi-modal fusion leveraging representations from both text and phoneme modalities, to reduce word error rate (WER) and perform robustly with varying input transcription quality. We demonstrate that PATCorrect consistently outperforms state-of-the-art NAR method on English corpus across different upstream ASR systems, with an overall 11.62\% WER reduction (WERR) compared to 9.46\% WERR achieved by other methods using text only modality. Besides, its inference latency is at tens of milliseconds, making it ideal for systems with low latency requirements. 

\end{abstract}

\section{Introduction}

Automatic speech recognition (ASR) model transcribes human speech into readable text, making it a critical component in large-scale natural language processing (NLP) systems like Amazon Alexa, Google Home and Apple Siri. Transcribed texts serve as input for downstream models such as intent detection in voice assistants and response generation in voice chatbots. Errors made in ASR transcriptions can severely impact the accuracy of downstream models and thus lower the performance of the entire NLP system. To improve the quality of ASR transcriptions, error correction models are applied to the outputs from ASR systems. ASR error correction can be formulated as a sequence-to-sequence (seq2seq) generation task, taking the ASR transcribed texts as input source sequences and the ground-truth transcriptions as target sequences. Previous studies \cite{d2016automatic, liao2020improving, mani2020asr} have proposed seq2seq models that decode the target sequence in an autoregressive (AR) manner. Although AR models have achieved great accuracy, high inference latency makes them infeasible for online production systems with low-latency constraints. For example, the end-to-end response time for voice digital assistants is of the order of tens of milliseconds for desired user experience. Under these latency constraints it is impractical to deploy AR models for error correction. Motivated by these considerations, we focus on non-autoregressive (NAR) models over AR models to meet the low latency requirement of industry production systems as NAR models are faster with parallel decoding during inference \cite{leng2021fastcorrect}. 

\begin{figure}[ht]
\centering
\includegraphics[width=230pt]{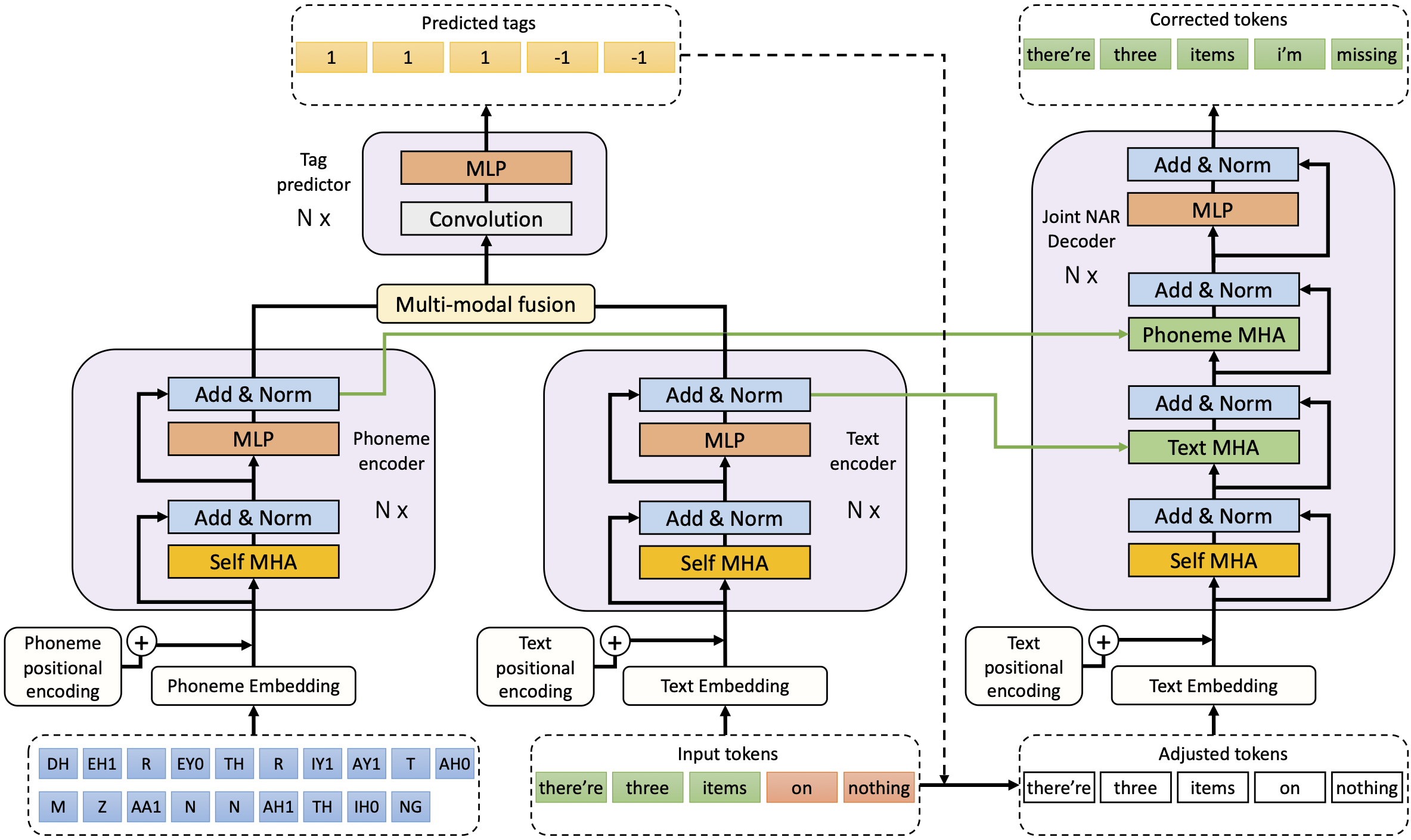}
\caption{Starting from the bottom, we input $\bm{w}$ and $\bm{p}$ to two encoders separately. The outputs from phoneme and text encoders are combined together by multi-modal fusion, and then fed into $TagP$ for adjusting source tokens. Two encoder-decoder attention layers are added sequentially in the joint NAR decoder for parallel decoding, to get the corrected target sequences.}
\label{fig1}
\end{figure}

We propose PATCorrect (Phoneme Augmented Transformer for ASR error Correction) as shown in Figure~\ref{fig1}, a novel NAR based upstream-model-agnostic ASR error correction model using both text and phoneme representations of the ASR transcribed sentences. PATCorrect applies multi-modal fusion to combine phoneme representation and text representation into joint feature embedding as input for the length tagging predictor. Both text and phoneme encoders interact with NAR decoder via encoder-decoder attention mechanism. PATCorrect improves the WER reduction (WERR) to 11.62\% compared to FastCorrect which is state-of-the-art (SOTA) NAR method that solely uses text only representation of the input, with comparable inference latency at the scale of tens of milliseconds. Our main contributions are summarized as follows:
\begin{itemize}
\item We propose PATCorrect, a novel model based on the Transformer architecture for NAR ASR correction. This model uses a multi-modal fusion approach that augments the traditional input text encoding with an additional phoneme encoder to incorporate pronunciation information, which is one of the key characteristics of spoken utterances.
\item Through extensive offline evaluations on English corpus, We demonstrate that PATCorrect outperforms the SOTA NAR ASR error correction model that uses text only modality. For example, PATCorrect improves WERR to 11.62\% with an inference latency at the same tens of milliseconds scale, while still being about 4.2 - 6.7x times faster than AR models.
\item To the best of our knowledge, we are the first to establish that multi-modal fusion is a promising direction for improving the accuracy of low latency NAR methods for ASR error correction, and comprehensively study the performance on English corpus across ASR systems with varying levels of quality.
\end{itemize}

\section{Related Work}
ASR error correction can be viewed as Neural Machine Translation (NMT) problem with erroneous sentences as source language, and corrected sentences as target language. \cite{cucu2013statistical,anantaram2018repairing} applied the NMT approach to domain-specific ASR systems for error correction, and further utilized ontology learning to repair ASR outputs by a 4-step method. Recent NMT methods based on Transformers \cite{vaswani2017attention, ng2019facebook} have become increasingly accurate and have inspired applications to ASR error correction \cite{liao2020improving,mani2020asr, hu2020deliberation}. Based on the intuition that phonetic information helps with understanding ASR errors \cite{fang2020using, sundararaman2021phoneme},  \cite{wang2020asr} found that adding phoneme information for domain-agnostic ASR system could benefit entity retrieval task. Although achieving high accuracy, these encoder-decoder based autoregressive generation models suffer from slow inference speed (hundreds of milliseconds), error propagation and demand for a large amount of training data.  

To address these issues in AR models, NAR sequence generation methods in NMT \cite{gu2017non}, which aim to speed up the inference of AR models while maintaining comparable accuracy, are gaining momentum in recent years. For NAR decoders, the length predictor is crucial as it outputs latent variables to determine the target sequence length for parallel generation. \cite{gu2019levenshtein} proposed dynamic insertion/deletion to iteratively refine the generated sequences based on previous predictions. \cite{ghazvininejad2019mask} used conditional masked language modeling for more efficient iterative parallel decoding. But straight-forward adaptation of these NAR methods from NMT to the ASR error correction problem may even lead to an increase in WER \cite{leng2021fastcorrect}. Thus current SOTA method for Chinese corpus error correction \cite{leng2021fastcorrect} chose to utilize edit alignment with text editing operations to train the length predictor and assign each source token with a length tag.
\section{PATCorrect for ASR Error Correction}
To correct the erroneous source tokenized sentence $\bm{w} = \{w_1, w_2, ..., w_n\}$ to the target error-free sequence $\bm{\hat{w}} = \{\hat{w_1}, \hat{w_2}, ..., \hat{w}_{\hat{n}} \}$, where the length of input tokens $n$ and the length of output tokens $\hat{n}$ can be the same or different, in PATCorrect we add the phoneme sequence $\bm{p} = \{p_1, p_2, ..., p_m\}$ of the source sentence $\bm{w}$ to represent the pronunciation information. During training, the text editing path from $\bm{w}$ to $\bm{\hat{w}}$ in the training set is pre-computed similar to \cite{leng2021fastcorrect}. Then both $\bm{w}$ and corresponding $\bm{p}$ are used as inputs to train a tag predictor which generates token-level alignment tags $\bm{t} = \{t_1, t_2, ..., t_n\}$. For each $w_i$, the corresponding $t_i$ consists of 4 possible edit operation types: $t_{i}=1$ means keep the token unchanged; $t_{i}=0$ means delete this token; $t_{i}=-1$ means substitute the token with another token; $t_{i}<-1$ means add other tokens adjacent to this token. Before input into the NAR decoder, $\bm{w}$ are adjusted based on the corresponding $\bm{t}$ in order to match $\bm{\hat{w}}$. During inference when we do not have ground truth target sequence to compute text editing paths, the tag predictor is used to predict the edit tags for the source sequence, and then the target sequence is generated according to predicted tags.
\subsection{Phoneme augmented encoder}
We augment the vanilla Transformer architecture \cite{vaswani2017attention} by adding an additional encoder to provide more information, which conceptually could encode any information modality depending on the application. Here we use phoneme sequence because homophone error is one of the major sources of ASR speech-to-text transcription errors \cite{fang2020using, sundararaman2021phoneme}. The two encoders first encode text and phoneme information separately without sharing model parameters. The stacked encoder layers, consisting of multi-head self-attention layers and position-wise fully connected feed-forward (MLPs) layers, transform the text sequence $\bm{w}$ and phoneme sequence $\bm{p}$ into hidden representations $H_{\bm{w}}=\{h_{w_1},h_{w_2},...h_{w_n}\} \in \mathbb{R}^{n \times d_h} $ and $H_{\bm{p}}=\{h_{p_1},h_{p_2},...h_{p_m}\} \in \mathbb{R}^{m \times d_h}$, with the output dimension $d_h$. We use the same hidden dimension $d_h$ for both encoders to simplify the multi-modal fusion operations later. There are two purposes for $H_{\bm{w}}$ and $H_{\bm{p}}$: firstly their fused representations are input into tag predictor to predict the edit alignment corresponding to each source token, secondly they both provide encoder-decoder cross attention in the joint NAR decoder during parallel decoding.
\subsection{Multi-modal fusion for tag predictor}
The tag predictor $TagP$ is trained using pre-computed ground truth tags with optimal editing alignment paths from source sequences to target sequences \cite{leng2021fastcorrect}. The encoder outputs $H_{\bm{w}}$ and $H_{\bm{p}}$ are combined into $H_{\bm{s}}$ via multi-modal fusion before feeding into the tag predictor to get $\bm{t} \in \mathbb{R}^{n \times 1}$, with the same length $n$ as the source sequence. We experiment with 3 different fusion approaches \cite{kiela2018efficient, ghazvininejad2018knowledge} and compare their performances in Sec.~\ref{sec:results}.

\noindent \textbf{Concatenation:} We concatenate the encoder outputs $H_{\bm{w}}$ and $H_{\bm{p}}$ so that $H_{\bm{s}}= \{h_{w_1},h_{w_2},...h_{w_n}, h_{p_1},h_{p_2},...h_{p_m}\} \in \mathbb{R}^{(n+m) \times d_h}$. Feeding this fused representation to convolutional and linear layers, the tag predictor $TagP$ will output one intermediate vector $I \in (n+m) \times 1$ that has the same length as the fused input $H_{\bm{s}}$. We then crop $I$ by selecting the first $n$ dimensions to get the final tag prediction $\bm{t} \in \mathbb{R}^{n \times 1}$.

\noindent \textbf{Pooling operations:} In this ASR error correction application where the length of phoneme sequences is longer than or equal to the length of the text sequences, \emph{i.e.} $m \geq n$, we pad $H_{\bm{w}} \in \mathbb{R}^{n \times d_h}$ with zeros to create $H'_{\bm{w}} \in \mathbb{R}^{m \times d_h}$. Then component-wise addition or max pooling can be applied to $H'_{\bm{w}}$ and $H_{\bm{p}}$:
\begin{equation}
   H_{\bm{s}} = \operatorname{Addition}(H'_{\bm{w}}, H_{\bm{p}}) \text{ or } H_{\bm{s}} = \operatorname{Max}(H'_{\bm{w}}, H_{\bm{p}}) 
\end{equation} 
where $H_{\bm{s}} \in  \mathbb{R}^{m \times d_h}$. 
Similarly, we crop the intermediate vector $I \in \mathbb{R}^{m \times 1}$ by selecting the first $n$ dimensions to get the final tag prediction $\bm{t} \in \mathbb{R}^{n \times 1}$.

\noindent \textbf{Cross attention:} We add cross attention layers with learnable parameters to project the phoneme encoder outputs on the text encoder outputs, which can handle the embedding length difference between the different modalities. We compute the cross attention outputs by taking $H_{\bm{w}}$ as query, $H_{\bm{p}}$ as key and value:
\begin{equation}
   H_{\bm{s}} = \operatorname{Softmax}(\frac{(H_{\bm{w}}W_c^Q)(H_{\bm{p}}W_c^K)^T}{\sqrt{d_h}})(H_{\bm{p}}W_c^V)
\end{equation}
where $W_c^Q, W_c^K, W_c^V \in \mathbb{R}^{d_h \times d_h}$ are the parameter matrices in cross attention layer with multiple attention heads, $H_{\bm{s}} \in \mathbb{R}^{n \times d_h}$ is the output after additional dropout, residual connection and normalization layers. No further cropping operation is needed since $TagP(H_{\bm{s}}) = \bm{t} \in \mathbb{R}^{n \times 1}$.

\subsection{Non-autoregressive joint decoder}
 Adapted from \cite{gu2017non}, our NAR model utilizes the tag predictor output as a latent variable to indicate the target length beforehand. Therefore, the training of PATCorrect is equivalent to optimizing the overall maximum likelihood of the conditional probability of $\bm{\hat{w}}$ with a variational lower bound:
\begin{multline}
\mathcal{L}_{\textrm{PATCorrect}}  = \log P_{\textrm{PATCorrect}}(\bm{\hat{w}}|\bm{w}, \bm{p}; \theta) \geq  \mathop{\mathbb{E}}_{\bm{t} \sim q} \\
\left( \sum_{i=1}^{n}\log P_{\textrm{TagP}}(t_i|\bm{w}, \bm{p}; \theta)
   + \sum_{i=1}^{\hat{n}}\log P(\hat{w}_i|\bm{w'}_{\bm{t}}, \bm{p}; \theta) \right) 
\end{multline}
where $\bm{w'}_{\bm{t}}$ is the adjusted source inputs based on $\bm{t}$, the actual input to the NAR decoder. With the optimal edit alignment paths, we provide an approximate distribution $q$ for the tag sequence with the most likely sample as the expectation. The first term within $\mathop{\mathbb{E}}_{\bm{t} \sim q}(\cdot)$ trains the tag predictor $TagP$, and the second term trains the error correction model. This design enables the parallel decoding for every target token $\hat{w}_i$ at the same time.

Meanwhile, we include the encoder-decoder attention mechanism for both encoders in the joint NAR decoder, by sequentially combining them together \cite{wang2020asr}. After getting the self-attention output from decoder $H_{\bm{\hat{w}}} \in \mathbb{R}^{\hat{n} \times d_h}$ with causal mask removed to enable parallel calculation 
 \cite{gu2017non}, we calculate the text and phoneme encoder-decoder attention by first using $H_{\bm{\hat{w}}}$ as query, text encoder output $H_{\bm{w}}$ as key and value:
\begin{equation}
   Z_w = \operatorname{Softmax}(\frac{(H_{\bm{\hat{w}}}W_w^Q)(H_{\bm{w}}W_w^K)^T}{\sqrt{d_h}})(H_{\bm{w}}W_w^V)
\end{equation}
and then using the output $Z_w$ as query, phoneme encoder output $H_{\bm{p}}$ as key and value:
\begin{equation}
   Z_p = \operatorname{Softmax}(\frac{(Z_wW_p^Q)(H_{\bm{p}}W_p^K)^T}{\sqrt{d_h}})(H_{\bm{p}}W_p^V)
\end{equation}
where $\{W_w^Q, W_w^K, W_w^V\}, \{W_p^Q, W_p^K, W_p^V\} \in \mathbb{R}^{d_h \times d_h}$ are the parameter matrices in the text and phoneme encoder-decoder multi-head attention layers, respectively. The remaining of the decoder layers are the same as the vanilla Transformer model where $Z_p \in \mathbb{R}^{\hat{n} \times d_h}$ will be input into fully-connected MLPs with residual connection and layer normalization.

\section{Experiments}
\label{sec:experiments}
\subsection{Datasets and ASR models}
We create the ASR transcription dataset $\mathbb{S}^{\textrm{trans}}$ specifically designed for correcting ASR errors. Multiple ASR systems are used to transcribe public English audio corpus, which has the corresponding ground truths from human transcriptions. The ASR transcriptions of the audio clips contain actual ASR errors, which are paired with golden texts. We combine two public English datasets, LibriSpeech \cite{panayotov2015librispeech} and Common Voice v9.0 \cite{ardila2019common}, together to get a large and diverse training corpus. In order to test different ASR systems with different architecture and performances, we choose 3 pre-trained ASR systems implemented in NeMo \cite{kuchaiev2019nemo}: The convolution-augmented Transformer \textbf{Conformer} \cite{gulati2020conformer} with top-tier performance; The convolution-based \textbf{Jasper} \cite{li2019jasper} with above average performance; A light-weight 5x5 \textbf{QuartzNet} \cite{kriman2020quartznet} with subpar performance. All 3 ASR systems are trained with the Connectionist Temporal Classification (CTC) loss. We use the default splits in LibriSpeech and Common Voice with transcriptions from 3 ASR systems together to compose the dataset $\mathbb{S}^{\textrm{trans}}$ as shown in Table~\ref{table1}, which in total has more than 3.5 million sentences pairs in training split. Note that LibriSpeech has been included in the training of all 3 ASR systems, so we use DEV and TEST splits from Common Voice as benchmarks in accuracy evaluations later.

To evaluate the performance of ASR error correction models, speed is measured by the latency of the whole inference process including encoding and decoding for different methods. For accuracy, similar to \cite{leng2021fastcorrect}, we use Word Error Rate (WER), WER Reduction (WERR), $Precision$ measures how many actual error tokens are edited among all of the edited tokens, $Recall$ measures how many actual error tokens are edited among all of the error tokens. Since unnecessary edits may even lead to WER increase, we use $F_{0.5}$ as an overall measurement to put more weight on $Precision$ for error detection ability \cite{liao2020improving,omelianchuk2020gector,rothe2021simple}. For error correction ability, $Correction$, is defined as the number of correctly edited error tokens divided by the number of edited error tokens, which measures the percentage of edited error tokens that match the ground truth.
\begin{table}[htb]
   \caption{ASR transcription dataset and ASR original WER}
   \label{table1}
   \begin{center}
   \begin{tabular}{cccc}
   \toprule
   Dataset $\mathbb{S}^{\textrm{trans}}$ & TRAIN & DEV & TEST  \\
   \midrule
   \# of sents &  1,171,348 & 21,898 & 21,877 \\
   Avg. words/sent & 15.8 & 12.1 & 11.8 \\
   \midrule
   Conformer WER & 4.56 & 7.20 & 7.28 \\
   Jasper WER & 7.80 & 14.12 & 15.29 \\
   QuartzNet WER & 18.97 & 32.96 & 35.86 \\
   \bottomrule 
   \end{tabular}
   \end{center}
\end{table}
\subsection{Model configurations} 
\textbf{PATCorrect}\hspace{0.3 cm} We use 6-layer text encoder, 6-layer phoneme encoder and 6-layer joint decoder with the hidden model dimension $d_h=512$, and MLP dimension $d_{\textrm{MLP}}=2048$. We use 8 attention heads for self-attention, encoder-decoder attention and cross attention respectively. In the cross attention setup for fusing two encoder outputs, we use 2 consecutive modules of cross attention with dropout, residual connection and layer normalization. For tag predictor $TagP$ trained with MSE loss, we apply 5 layers of convolutional modules which consists of 1-D convolution layer with kernel size of 3, ReLU, layer normalization and dropout, followed by 2 layers of MLPs.

\noindent \textbf{Baseline models}\hspace{0.3 cm} We compare our model with both AR Transformer and popular NAR methods. For \textbf{AR Transformer}, we use the standard vanilla Transformer architecture with 6-layer encoder and 6-layer decoder and hidden size $d_h=512$. For NAR methods Mask Predict (\textbf{CMLM}) \cite{ghazvininejad2019mask} and Levenshtein Transformer (\textbf{LevT}) \cite{gu2019levenshtein}, we use the default training hyperparameters. For SOTA NAR ASR error correction method \textbf{FastCorrect} \cite{leng2021fastcorrect},
we use the same 6-layer encoder-decoder architecture and the same architecture for length predictor.

\noindent \textbf{Training and inference details}\hspace{0.3 cm} All models are implemented using Fairseq \cite{ott2019fairseq}, and trained using 4 NVIDIA Tesla V100 GPUs with maximum batch token size of 5000 and label smoothed cross entropy loss function. They are trained from scratch using the ASR transcription dataset $\mathbb{S}^{\textrm{trans}}$ for 30 epochs. Source and target sentences are tokenized using sentencepiece \cite{kudo2018sentencepiece}, and the phoneme sequences are generated by English grapheme to phoneme conversion using the CMU pronouncing dictionary \cite{g2pE2019}, an independent post-processing step of the 1-best ASR hypothesis as model inputs. We use Adam optimizer \cite{kingma2014adam} and inverse square root for learning rate scheduling starting from $5e^{-4}$. During inference, we set the test batch size as 1 to simulate the online production environment with output from the upstream ASR system, and all NAR methods have the same max decoding iteration of 1. Different hardware conditions are tested including single NVIDIA Tesla V100 GPU, 8 CPUs and 4 CPUs with Intel(R) Xeon(R) CPU E5-2686 v4 @ 2.30GHz.

\section{Results}
\label{sec:results}
\subsection{Accuracy}
We compare the WER and WERR using different error correction models for all three ASR systems and their total combined transcriptions on Common Voice TEST split, as shown in Table~\ref{table2}. PATCorrect outperforms other NAR methods for all three ASR system transcriptions. Results show: 1) For averaged total results on three ASR systems, PATCorrect beats the SOTA FastCorrect method by improving the TEST set WERR from 9.46 to 11.62, which is more than 20\% relative improvement; 2) Among all of the multi-modal fusion operations experimented, cross attention performs best across almost all datasets.
\begin{table}[htb]
   \caption{
   WER (\%) $\downarrow$ and WERR (\%) $\uparrow$ using error correction models averaged over outputs from 3 upstream ASR systems}
   \label{table2}
   \centering
   \begin{tabular}{c|cc|cc}
   \toprule
   \multirow{2}{*}{Models} & \multicolumn{2}{c|}{WER $\downarrow$} & \multicolumn{2}{c}{WERR $\uparrow$} \\
      & DEV & TEST & DEV & TEST \\
   \midrule
   No error correction & 25.55 & 27.96 & - & -\\
   AR Transformer & 19.45 & 22.82 & 23.87 & 18.40 \\
   \midrule
   CMLM & 23.92 & 26.49 & 6.36 & 5.26 \\
   LevT & 23.19 & 25.70 & 9.24 & 8.11 \\
   FastCorrect & 22.13 & 25.32 & 13.37 & 9.46 \\
   PATCorrect$(cat)$ & 22.09 & 25.24 & 13.53 & 9.75\\
   PATCorrect$(add)$ & 22.14 & 25.28 & 13.33 & 9.61 \\
   PATCorrect$(max)$ & 21.93 & 25.07& 14.16 & 10.35 \\
   PATCorrect$(cross\_atten)$ & \textbf{21.57} & \textbf{24.72} & \textbf{15.59} & \textbf{11.62}\\
   \bottomrule
   \end{tabular}
   \end{table}
\subsection{Speed}
We test inference speed on both GPU and CPUs as shown in Table~\ref{table3} for Common Voice TEST split. Consistent with the observation from \cite{gu2017non}, AR model shows a linear latency trend with decoding lengths, while the latency of the NAR methods increases slightly. PATCorrect achieves an inference latency comparable with other NAR models, especially on GPU hardware, while still being about 4.2 - 6.7x times faster than AR models.
\begin{table}[htb]
   \caption{
   Inference latency $\downarrow$ with unit ms/sentence}
   \label{table3}
   \centering
   \begin{tabular}{c|ccc}
    \toprule
   Models & 1$\times$ GPU & 8$\times$CPUs & 4$\times$CPUs \\
   \midrule
   AR Transformer & 141.00 & 149.79 & 186.89 \\
   \midrule
   CMLM & 35.54 & 43.78 & 47.90 \\
   LevT & 47.86 & 45.68 & 55.33 \\
   FastCorrect & \textbf{20.81}  & \textbf{23.77} & \textbf{29.14} \\
   PATCorrect$(cross\_atten)$ & 33.71 & 41.89 & 49.97 \\
   \bottomrule
   \end{tabular}
   \end{table}
\subsection{Sensitivity analysis}
 We conduct sensitivity analysis by comparing the error detection ability and error correction ability for different models with 3 ASR transcriptions on Common Voice TEST split. In Table~\ref{table4}, $P, R, C$ denote $Precision, Recall, Correction$ respectively. Results show that our PATCorrect not only has the highest $F_{0.5}$ score with great $Precision$ that is comparable to AR model, but also has better ability to edit error tokens to the correct targets indicated by higher $Correction$.
\begin{table}[ht]
  \caption{
  Sensitivity metrics $\uparrow$ for error correction models}
  \label{table4}
  \centering
  \begin{tabular}{c|cccc}
  \toprule
    Models  & $P$ & $R$ & $F_{0.5}$ & $C$ \\
  \midrule
  AR Transformer& 90.89 & 65.33 & 84.30 & 37.81 \\
  \midrule
  CMLM & 84.87 & 61.45 & 78.86 & 27.37 \\
  LevT & 86.49 & \textbf{61.75} & 80.07 & 28.42 \\
  FastCorrect & 89.53 & 60.90 & 81.83 & 27.70 \\
  PATCorrect $(cross\_atten)$ & \textbf{90.27} & 59.64 & \textbf{81.86} & \textbf{29.50} \\
  \bottomrule
  \end{tabular}
  \end{table}   
\subsection{Ablation study}
We perform ablation study to understand the effectiveness of different components of PATCorrect by removing a component while retaining the others. No phoneme input for $TagP$ means that we only use text information as input for predicting token tags while still using phoneme encoder-decoder attention in NAR decoder. No phoneme attention means that we remove phoneme encoder-decoder attention from the NAR decoder and only use text encoder-decoder attention, while still using cross-attention to fuse the text and phoneme encoder outputs for $TagP$. We also increase the amount of pre-training data by using a synthetic dataset $\mathbb{S}^{\textrm{synth}}$ for data augmentation to pretrain the model for 20 epochs. We crawl 50M sentences from English Wiki pages \cite{Wikiextractor2015}, and add random editing operations like deletion, insertion, substitution with a homophone dictionary to produce erroneous sentences paired with the original correct texts, that mimics the ASR errors with a simulated error distribution. The results in Table~\ref{table5} with WER and WERR on the Common Voice TEST split, equal-weight averaged from all 3 ASR systems, show that adding phoneme information in $TagP$ and NAR decoder lead to better WER and WERR. In addition, using $\mathbb{S}^{\textrm{synth}}$ for pretraining can also boost the model accuracy to further reduce WER.
\begin{table}[htb]
   \caption{
   Ablation study for PATCorrect model}
   \label{table5}
   \centering
   \begin{tabular}{c|cc}
   \toprule
     Models &  WER $\downarrow$ & WERR $\uparrow$\\
   \midrule
   No error correction & 27.96 & - \\
   FastCorrect & 25.32 & 9.46 \\
   \midrule
   No phoneme input for $TagP$  & 25.13 & 10.14 \\
   No phoneme attention in decoder & 25.06 & 10.39 \\
   PATCorrect$(cross\_atten)$ & 24.72 & 11.62 \\
   PATCorrect$(cross\_atten)$ + $\mathbb{S}^{\textrm{synth}}$  & \textbf{24.33} & \textbf{13.00}\\
   \bottomrule
   \end{tabular}
   \end{table}

\section{Conclusions}
We propose PATCorrect, a novel NAR phoneme-augmented Transformer-based model with robust performance on different upstream ASR systems with varying speech-to-text transcription quality. Our model outperforms SOTA NAR ASR error correction models, and still 4.2 - 6.7x times faster than AR models, which makes it a great fit as industrial scale text editing method to refine ASR transcriptions. Our study establishes that multi-modal fusion is a promising direction for improving the accuracy of low latency NAR methods for ASR error correction.


\newpage
\bibliographystyle{IEEEtran}
\bibliography{PATCorrect}

\end{document}